\documentclass{pprai}

\usepackage[utf8]{inputenc}
\usepackage[T1]{fontenc}
\usepackage{graphicx}
\usepackage{amsthm}
\usepackage{txfonts}
\usepackage{url}
\usepackage{algpseudocode}

\usepackage{todonotes}      

\title{LlamBERT: Large-scale low-cost data annotation in NLP}
\headtitle{LlamBERT: Large-scale low-cost data annotation in NLP}

\usepackage{subcaption}
\usepackage{svg}
\usepackage{booktabs}       
\usepackage{multirow}
\usepackage{hhline}
\usepackage[super]{nth}
\usepackage{hyperref}

\author{Bálint Csanády$^1$, Lajos Muzsai$^1$, Péter Vedres$^1$\\
Zoltán Nádasdy$^{2,3}$, András Lukács$^1$} 
\headauthor{B. Csanády, L. Muzsai, P. Vedres, Z. Nádasdy, A. Lukács}
\affiliation{
  $^1$ELTE Eötvös Loránd University, 
  Institute of Mathematics,\\ AI Research Group\\
  csbalint@protonmail.ch, muzsailajos@protonmail.com,\\  vedrespeter0000@gmail.com,
  andras.lukacs@ttk.elte.hu\\
  $^2$ELTE Eötvös Loránd University, Institute of Psychology\\
  $^3$The University of Texas at Austin, Department of Psychology \\
  zoltan@utexas.edu \\
  }


\begin{document}
\maketitle


\begin{abstract}
Large Language Models (LLMs), such as GPT-4 and Llama\,2, show remarkable proficiency in a wide range of natural language processing (NLP) tasks.
Despite their effectiveness, the high costs associated with their use pose a challenge.
We present LlamBERT, a hybrid approach that leverages LLMs to annotate a small subset of large, unlabeled databases and uses the results for fine-tuning transformer encoders like BERT and RoBERTa.
This strategy is evaluated on two diverse datasets: the IMDb review dataset and the UMLS Meta-Thesaurus.
Our results indicate that the LlamBERT approach slightly compromises on accuracy while offering much greater cost-effectiveness.
\end{abstract}


\section{Introduction}
In the contemporary technological landscape, when confronted with the task of annotating a large corpus of natural language data using a natural language prompt, LLMs such as the proprietary GPT-4 \cite{gpt4} and the open-source Llama\,2 \cite{Llama2} present themselves as compelling solutions.
Indeed, minimal prompt-tuning enables them to be highly proficient in handling a wide variety of NLP tasks \cite{gpteval}.
However, running such LLMs on millions of prompts demands large and expensive computational resources.
There have been optimization efforts aimed at achieving superior performance with reduced resource requirements \cite{selfinstruct,gptq}.
Numerous studies have investigated the efficiency and resource requirements of LLMs 
versus smaller transformer encoders and humans 
\cite{su2022selective,yu2023open,2023Gilardi,savelka2023unreasonable,alizadeh2023open,sprenkamp2023large}.
Recent advancements in data augmentation with LLMs \cite{ding2024data} underscore our approach, which relies on data labeling.
Going beyond the exclusive use of LLMs for a task, we combine LLMs with substantially smaller yet capable NLP models.
A study closest to our approach is \cite{frei2023annotated}, where GPT-NeoX was used to surrogate human annotation for solving named entity recognition.

Through two case studies, our research aims to assess the advantages and limitations of the approach we call LlamBERT, a hybrid methodology utilizing both LLMs and smaller-scale transformer encoders.
The first case study examines the partially annotated IMDb review dataset \cite{IMDb} as a comparative baseline, while the second selects biomedical concepts from the UMLS Meta-Thesaurus \cite{UMLS} to demonstrate potential applications.
Leveraging LLM's language modeling capabilities, while utilizing relatively modest resources, enhances their accessibility and enables new business opportunities.
We believe that such resource-efficient solutions can foster sustainable development and environmental stewardship.

    
\section{Approach}
Given a large corpus of unlabeled natural language data, the suggested LlamBERT approach takes the following steps:
(i) Annotate a reasonably sized, randomly selected subset of the corpus utilizing Llama\,2 and a natural language prompt reflecting the labeling criteria;
(ii) Parse the Llama\,2 responses into the desired categories; 
(iii) Discard any data that fails to classify into any of the specified categories;
(iv) Employ the resulting labels to perform supervised fine-tuning on a BERT classifier;
(v) Apply the fine-tuned BERT classifier to annotate the original unlabeled corpus.

We explored two binary classification tasks, engineering the prompt to limit the LLM responses to one of the two binary choices.
As anticipated, our efforts to craft such a prompt were considerably more effective when utilizing the 'chat' variants of Llama\,2 \cite{RLHF}.
We investigated two versions: {\small\texttt{Llama-2-7b-chat}} 
running on a single A100 80GB GPU, and 
{\small\texttt{Llama-2-70b-chat}} requiring four such GPUs.
We also tested the performance of {\small\texttt{gpt-4-0613}} using the OpenAI API.

\section{The IMDb dataset}
The Stanford Large Movie Review Dataset (IMDb) \cite{IMDb} is a binary sentiment dataset commonly referenced in academic literature.
It comprises 25,000 labeled movie reviews for training purposes, 25,000 labeled reviews designated for testing, and an additional 50,000 unlabeled reviews that can be employed for supplementary self-supervised training.
This dataset serves as a fundamental baseline in NLP for classification problems, which allows us to evaluate our method against a well-established standard \cite{top1_IMDb, top2_IMDb, top3_IMDb}.

\subsection{Experimental results}
All of the results in this section were measured on the entire IMDb sentiment test data.
In Table \ref{tableIMDb_Llama}, we compare the performance of Llama\,2 and GPT-4 in different few-shot settings.
Due to limited access to the OpenAI API, we only measured the 0-shot performance of GPT-4.
The results indicate that the number of few-shot examples has a significant impact on {\small\texttt{Llama-2-7b-chat}}.
This model exhibited a bias toward classifying the reviews as positive, but few-shot examples of negative sentiment effectively mitigated this.
Likely due to reaching the context-length limit, 3-shot prompts did not outperform 2-shot prompts on {\small\texttt{Llama-2-7b-chat}}, achieving an accuracy of 87.27\%.
The inference times shown in Table \ref{tableIMDb_Llama} depend on various factors, including the implementation and available hardware resources; they reflect the specific setup we used at the time of writing.

\begin{table}[!h]
    \caption{Comparison LLM test performances on the IMDb data.}\label{tableIMDb_Llama}
    \centering
    \begin{tabular}{|l|c|c|c|c|c|c|}
    \hline
                    & \multicolumn{3}{c|}{\textbf{Accuracy \%}} & \multicolumn{3}{c|}{\textbf{Inference time}}  \\ 
    \cline{2-7}
    \textbf{LLM}    & \textbf{0-shot} & \textbf{1-shot} & \textbf{2-shot} & \textbf{0-shot} & \textbf{1-shot} & \textbf{2-shot} \\ 
    \hline\hline
    Llama-2-7b-chat & 75.28 & 89.77 & 93.93 & 3h 54m  & 4h 16m & 8h 14m \\ 
    Llama-2-70b-chat& 95.39 & 95.33 & 95.42 & 28h 11m     & 39h 6m & 76h 2m \\ 
    gpt-4-0613      & 96.40 & N/A   & N/A   & 49h 11m & N/A & N/A \\ 
    \hline
    \end{tabular}
\end{table}

In Table \ref{tableIMDb}, we compare various pre-trained BERT models that were fine-tuned for five epochs on different training data with a batch size of 16.
First, we established a baseline by using the original gold-standard training data.
For the LlamBERT results, training data labeling was conducted by the {\small\texttt{Llama-2-70b-chat}} model from 0-shot prompts.
The LlamBERT results were not far behind the baseline measurements, underscoring the practicality and effectiveness of the framework.
Incorporating the extra 50,000 unlabeled data in LlamBERT resulted in a slight improvement in accuracy.
We also evaluated a combined strategy where we first fine-tuned with the extra data labeled by {\small\texttt{Llama-2-70b-chat}}, then with the gold training data.
The large version of RoBERTa performed the best on all 4 training scenarios, reaching a state-of-the-art accuracy of 96.68\%.
Inference on the test data with {\small\texttt{roberta-large}} took 9m 18s, after fine-tuning for 2h 33m.
Thus, we can estimate that labeling the entirety of IMDb's 7.816 million movie reviews \cite{IMDb_stat} would take about 48h 28m with {\small\texttt{roberta-large}}.
In contrast, the same task would require approximately 367 days on our setup using {\small\texttt{Llama-2-70b-chat}}, while demanding significantly more computing power.

\begin{table}[!h]
    \caption{Comparison BERT test accuracies on the IMDb data.}
    \label{tableIMDb} 
    \centering
    \begin{tabular}{|l|c|c|c|c|}
    \hline
    \textbf{BERT} & \textbf{Baseline} & \textbf{LlamBERT} & \textbf{LlamBERT}     & \textbf{Combined} \\
    \textbf{model}    & \textbf{train}    & \textbf{train}    & \textbf{train\&extra} & \textbf{extra+train} \\
    \hline\hline
    distilbert-base \cite{distilbert} & 91.23 & 90.77 & 92.12 & \textbf{92.53} \\ 
    bert-base       & 92.35 & 91.58 & 92.76 & \textbf{93.47} \\ 
    bert-large      & 94.29 & 93.31 & 94.07 & \textbf{95.03} \\ 
    roberta-base    & 94.74 & 93.53 & 94.28 & \textbf{95.23} \\ 
    roberta-large    & 96.54 & 94.83 & 94.98 & \textbf{96.68} \\ 
    \hline
    \end{tabular}
\end{table}

\subsection{Error analysis}

\begin{figure}[!h]
     \centering
     \hspace*{-1cm}
     \begin{subfigure}[b]{0.55\textwidth}
        \centering
        \includegraphics[width=\textwidth]{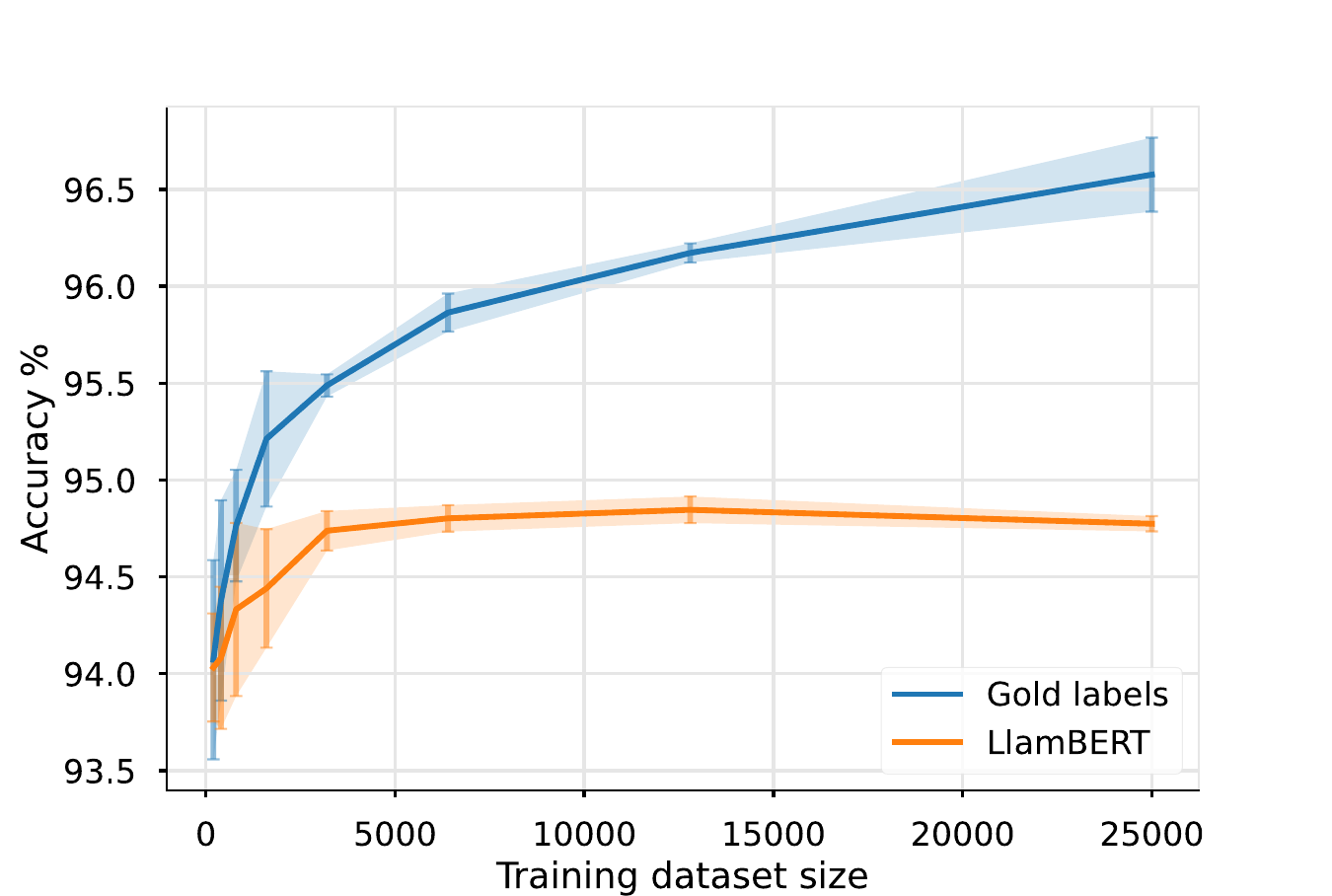}
    \end{subfigure}
    \hspace*{-0.7cm}
    \begin{subfigure}[b]{0.55\textwidth}
        \centering
        \includegraphics[width=\textwidth]{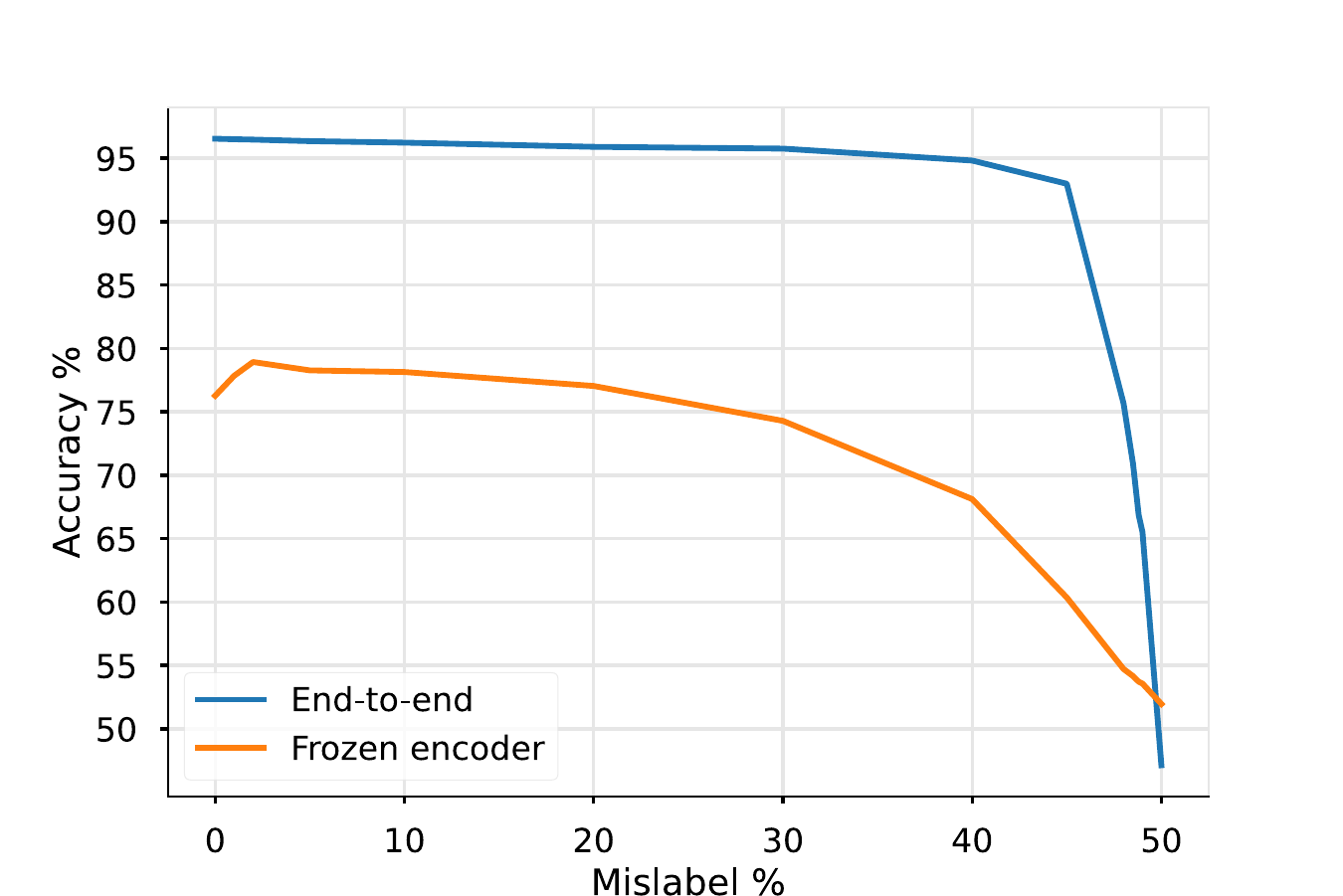}
    \end{subfigure}
    \hspace*{-1.50cm}
    \caption{
    Accuracy (\%) comparison of RoBERTa classifiers on the IMDb test data.
    On the left: The effects of training data size. 
    On the right: The effects of intentionally mislabeling a random part of the gold training data.}
     \label{figIMDb}
\end{figure}

To assess the relationship between training data quantity and the accuracy of the ensuing BERT model,
we fine-tuned {\small\texttt{roberta-large}} across different-sized subsets of the gold training data as well as data labeled by {\small\texttt{Llama-2-70b-chat}}.
As the left side of Fig.\,\ref{figIMDb} indicates, the performance improvement attributed to the increasing amount of training data tends to plateau more rapidly in the case of LlamBERT.
Based on these results, we concluded that labeling 10,000 entries represents a reasonable balance between accuracy and efficiency for the LlamBERT experiments in the next section.
We were also interested in assessing the impact of deliberately mislabeling various-sized random subsets of the gold labels.
The discrepancy between the gold-standard training labels and those generated by Llama\,2 stands at 4.61\%; this prompted our curiosity regarding how this 4.61\% error rate compares to mislabeling a randomly chosen subset of the gold training data.
As shown on the right side of Fig.\,\ref{figIMDb}, {\small\texttt{roberta-large}} demonstrates substantial resilience to random mislabeling.
Furthermore, data mislabeled by {\small\texttt{Llama-2-70b-chat}} results in a more pronounced decrease in performance compared to that of a random sample.


\begin{table}[!h]
    \caption{Comparison of human annotation to model outputs on wrong test answers.}
    \label{tableIMDbErr} 
    \centering
    \begin{tabular}{|l||c|c|c||c|c|c|}
    \hline
    RoBERTa   & \multicolumn{3}{c||}{\textbf{LlamBERT train}}  & \multicolumn{3}{c|}{\textbf{Combined extra+train}} \\
    \cline{2-7}
    sentiment & positive & negative & mixed & positive & negative & mixed \\
    \hline
    \hline
    positive  & 31       & 16       & 13    & 25       & 17       & 13 \\ 
    negative  & 17       & 14       & 9     & 15       & 14       & 16 \\ 
    \hline
    \end{tabular}
\end{table}
\vspace*{-.8cm}

We also conducted a manual error analysis on two of the models fine-tuned from {\small\texttt{roberta-large}}.
For the model fine-tuned with the combined strategy, we randomly selected 100 reviews from the test data, where the model outputs differed from the gold labels.
We sampled an additional 27 mislabeled reviews of the model fine-tuned with the LlamBERT strategy to get a sample size of 100 on the errors of this model too.
We collected human annotations for the sentiment of the selected reviews independently from the gold labels.
In the case of human annotation, we added a third category of \emph{mixed/neutral}.
Reviews not discussing the movie or indicating that 'the film is so bad it is good' were typically classified in this third category.
Table \ref{tableIMDbErr} compares the human annotations to the model outputs. 
The results indicate a comparable ratio of positive to negative labels between the human annotations and the model outputs, suggesting that the model outputs are more aligned with human sentiment than the original labels.
Overall human performance on this hard subset of the test data was worse than random labeling.

\section{The UMLS dataset}
The United Medical Language System (UMLS) \cite{UMLS}, developed by the United States National Library of Medicine, 
is a comprehensive and unified collection of nearly 200 biomedical vocabularies.
It has played a crucial role in fields such as natural language processing, ontology development, and information retrieval for over 30 years \cite{UMLSdevelopment}.
The UMLS Metathesaurus consolidates various lexical variations of terms into single concepts, outlining their interrelationships.
However, its breadth, with over 3 million concepts, complicates the selection of specific subsets for research due to its
vague semantic labels.
Faced with the need to identify a distinct subset of the 
Metathesaurus for subsequent research, we aimed to classify anatomical entities within it,
based on their relevance to the human nervous system.
Previous research on creating a neurological examination ontology involved extracting terms from case studies and manually mapping them to UMLS concepts, a task that can be extremely labor-intensive \cite{hier2020neuro}.
Our approach streamlines this process by efficiently leveraging the vast amount of knowledge condensed into LLMs and mitigates the need for expert annotation.

By selecting relevant semantic types spanning multiple biological scales, but excluding genes, we were able to reduce the number of concepts to approximately 150,000 anatomical structures, resulting in a still substantially large dataset.
Among these anatomical structures, we sought to find concepts related to the human nervous system, excluding purely vascular or musculoskeletal structures, and indirectly related entities such as the outer ear and eye lens.
Using distinct random samples, we annotated 1,000 concepts for testing and an additional 1,000 
for hand-labeled fine-tuning.
We opted for a 1-shot prompt, on which {\small\texttt{Llama-2-7b-chat}} achieved an accuracy of 87.5\%, while {\small\texttt{Llama-2-70b-chat}} reached 96.5\%, and {\small\texttt{gpt-4-0613}} scored 94.6\%.
For fine-tuning BERT models, we labeled a distinct set of 10,000 concepts with {\small\texttt{Llama-2-70b-chat}}.

\subsection{Experimental results}

As shown in Table \ref{tableUMLS_reg}, fine-tuning general BERT models on the baseline hand-labeled 
dataset already yielded commendable results
, however, our LlamBERT approach further improved these outcomes. 
Moreover, the combined strategy marginally surpassed Llama\,2's initial performance.
Within the biomedical domain, specific BERT models such as {\small\texttt{BiomedBERT-large}} \cite{biomedbert} were already accessible and predictably outperformed both {\small\texttt{bert-large}} and {\small\texttt{roberta-large}} across all training scenarios.
Yet, the combined approach using {\small\texttt{roberta-large}} demonstrated comparable performance, suggesting that our methodology could serve as an alternative to training domain-specific models.

\begin{table}[!h]
    \caption{
    Accuracy comparison of different training data for the UMLS classification; \nth{95} percentile confidence interval measured on 5 different random seeds.
    }\label{tableUMLS_reg}
    \centering
    \begin{tabular}{|l|c|c|c|}
    \hline
    \textbf{Model} & \textbf{Baseline} & \textbf{LlamBERT} & \textbf{Combined} \\ 
    \hline\hline
    bert-large       & 94.84 ($\pm$0.25) & 95.70 ($\pm$0.21) & 96.14 ($\pm$0.42)\\  
    roberta-large    & 95.00 ($\pm$0.18) & 96.02 ($\pm$0.12) & 96.64 ($\pm$0.14)\\ 
    BiomedBERT-large & 96.72 ($\pm$0.17) & 96.66 ($\pm$0.13) & 96.92 ($\pm$0.10)\\  
    \hline
    \end{tabular}
\end{table}

\section{Conclusions}
Through two case studies showcasing the LlamBERT technique, we demonstrated the feasibility of efficiently labeling large quantities of natural language data with state-of-the-art LLMs.
Combining the LlamBERT technique with fine-tuning on gold-standard data yielded the best results in both cases, achieving state-of-the-art accuracy on the IMDb benchmark.
Our code is available on GitHub\footnote{\url{https://github.com/aielte-research/LlamBERT}}.

To further increase the quality of data initially provided by the LLM annotation, we aim to incorporate PEFT \cite{PEFT} techniques such as LoRA \cite{LoRA}, prefix tuning \cite{prefix_tuning}, and P-tuning \cite{P-tuning} in the future.

\section*{Acknowledgments}
The authors 
thank the support of the National Research, Development and Innovation Office within the framework of the Thematic Excellence Program 2021 -- National Research Sub programme: “Artificial intelligence, large networks, data security: mathematical foundation and applications" and the Artificial Intelligence National Laboratory Program (MILAB). 
We appreciate the support provided by OpenAI under the Researcher Access Program.
We thank Máté Márk Horváth and Virág Bálint for their assistance in labeling the UMLS test dataset.


\bibliography{pprai}
\bibliographystyle{pprai}

\newpage
\section{Appendix}
This appendix outlines the two prompts we used to engage the Llama 2 model for our article’s case studies.
Few-shot examples contained the same prompt structure continued by the appropriate answer. 
\subsection{IMDB prompt}

\begin{verbatim}
[INST] <<SYS>>
Please answer with 'positive' or 'negative' only!
<</SYS>>
Decide if the following movie review is positive or negative: 
<text of the review>
If the movie review is positive please answer 'positive',
if the movie review is negative please answer 'negative'.
Make your decision based on the whole text.
[/INST]
\end{verbatim}

\subsection{UMLS prompt}
\begin{verbatim}
[INST] <<SYS>>
Please answer with a 'yes' or a 'no' only!
<</SYS>>
Decide if the term: <available synonyms of the term separated by a ;>
is related to the human nervous system.
Exclude the only vascular structures,
even if connected to the nervous system.
If multiple examples or terms with multiple words are given,
treat them all as a whole and make your decision based on that.
[/INST]
\end{verbatim}
\end{document}